\documentclass[sigconf]{acmart}
\setcopyright{none}
\renewcommand\footnotetextcopyrightpermission[1]{}
\acmConference[arXiv]{XXX}{December 2022}{Berkeley, CA, USA}

\begin{document}

\title{Reinforcement Learning for Resilient Power Grids}

\author{Zhenting Zhao}
\affiliation{%
  \institution{University of California-Berkeley}
  \city{Berkeley}
  \state{California}
  \country{USA}}
 \email{zzalex@berkeley.edu}
 
 \author{Po-Yen Chen}
\affiliation{%
  \institution{University of California-Berkeley}
  \city{Berkeley}
  \state{California}
  \country{USA}}
 \email{andy\_chen@berkeley.edu}
 
 \author{Yucheng Jin}
\affiliation{%
  \institution{University of California-Berkeley}
  \city{Berkeley}
  \state{California}
  \country{USA}}
 \email{yuchengjin@berkeley.edu}

\renewcommand{\shortauthors}{Zhenting Zhao et al.}

\begin{abstract}

Traditional power grid systems have become obsolete under more frequent and extreme natural disasters. Reinforcement learning (RL) has been a promising solution for resilience given its successful history of power grid control. However, most power grid simulators and RL interfaces do not support simulation of power grid under large-scale blackouts or when the network is divided into sub-networks. In this study, we proposed an updated power grid simulator built on Grid2Op, an existing simulator and RL interface, and experimented on limiting the action and observation spaces of Grid2Op. By testing with DDQN and SliceRDQN algorithms, we found that reduced action spaces significantly improve training performance and efficiency. In addition, we investigated a low-rank neural network regularization method for deep Q-learning, one of the most widely used RL algorithms, in this power grid control scenario. As a result, the experiment demonstrated that in the power grid simulation environment, adopting this method will significantly increase the performance of RL agents.

\end{abstract}



\maketitle

\section{Introduction}
\subsection{Investigation into reinforcement learning on power grids}

How to stabilize the functionality and operation of electrical grids under extreme conditions is an important topic in the field of electrical engineering, since the unexpected malfunction or breakdown of electrical grids might cause significant loss of life and property. A well-known example that demonstrates the horrendous consequences of electrical grids failure is 2021 Texas power crisis that left more than 10 million people without electricity \cite{busby2021cascading}. According to official reports, 2021 Texas power crisis caused more than 210 casualties \cite{ulrich2022texas} and about \$200 billion economic loss \cite{wu2021open}.

Smart grid is the most promising technology to be the solution for power outage. Smart grid is highly autonomous, efficient in power distribution, and resilient under extreme situations \cite{tuballa2016review}. In the past years, many reinforcement learning (RL)-based smart grids were proposed by researchers \cite{zhang2018review}, their applications include energy management, demand response, operational control, etc. \cite{zhang2019deep}. 

We conducted a comparative study of two existing RL algorithms, double deep Q-network (DDQN) and slice recurrent deep Q-network (SliceRDQN), on smart grids with reduced action and observation spaces and proposed an applicable algorithm with low-rank regularization (LRR) to improve resilience. 

We used Grid2Op framework for simulation and chose two scenarios for comparative experiments: a simpler case with only 5 substations in the electrical grid (rte\_case5\_example) and a more complicated and realistic case with 14 substations (rte\_case14\_realistic). We selected three metrics for algorithm performance evaluation: time-based metrics, event-based metrics, and economic-based metrics.

The rest of this paper is organized as follows. Section 1.2 introduces low-rank regularization on deep Q-learning. Section 2 summarizes related studies on smart grid, reinforcement learning, and low-rank regularization. Section 3 explains the simulation environment, action spaces, observation spaces, and evaluation metrics in detail. Section 4 describes the experimental setup and discusses the experimental results. Finally, Section 5 serves as the conclusion of this paper.

\subsection{Low-rank regularization on deep Q-learning}

Deep Q-learning (DQN) is a fundamental methodology for solving a Markov decision process (MDP) problem. The value function, $Q(s, a)$, is defined as the expected value when an agent starts in state $s$, takes action $a$, and then follows a sequence of actions. Mathematically, it obeys the Bellman recursion,

\begin{equation} \label{eq:define-q}
Q(s,a) = r(s,a) + \mathbf{E}_{s' \sim P(s'|s,a)} \left [\gamma \max_{a'\in \mathcal{A}} Q(s',a') \right]
\end{equation}

Where $\mathcal{A}$ is the action space, $r$ is the reward function, and $P$ is the probabilistic transition function which gives the probability of transitioning into state $s'$ from taking action $a$ at the current state $s$.

The goal of deep Q-learning is to train a deep neural network to represent $Q(s,a)$ and obtain policy $\pi(s) = argmax_{a\in \mathcal{A}} Q(s,a)$. During training, the model samples $|\mathcal{B}|$ times and collects  

$\{(s_{t}^{(i)}, r_{t}^{(i)}, a_{t}^{(i)}, s_{t+1}^{(i)})\}_{i=1}^{|\mathcal{B}|}$, and forms the following updating targets,

\begin{equation} \label{eq:dqn-y}
   y^{(i)}=r_{t}^{(i)}+\gamma \max_{a' \in \mathcal{A}} Q(s_{t+1}^{(i)},a'; \theta)
\end{equation}

Where $Q(s_{t+1}^{(i)},a'; \theta)$ is the learned neural network, named as the value network, to represent the value function $Q(s,a)$ where $\theta$ denotes the parameters of the neural network which characterizes it. The value network is then updated by taking a gradient step for the loss function,

\begin{equation} \label{eq:dqn-loss}
\sum_{i=1}^{|\mathcal{B}|}(y^{(i)}-Q(s_{t}^{(i)},a_{t}^{(i)}; \theta))^2
\end{equation}

Many works in RL and control theory show the value function $Q(s,a)$ has some low-rank property in a variety of tasks (\cite{DBLP:journals/corr/Ong15}, \cite{DBLP:conf/cdc/AloraGKML16}, \cite{DBLP:conf/iclr/YangZXK20}). For example, \cite{DBLP:journals/corr/Ong15} shows that, if we denote the value function in a matrix form, $Q \in \mathbb{R}^{|\mathcal{S}| \times |\mathcal{A}|}$, called the Q table, then matrix $Q$ can be decomposed as $Q = L + E$, where $L$ is a low-rank matrix and $E$ is a sparse matrix. Intuitively, this means the Q table is composed mainly of the low-rank component $L$ with a sparse noise $E$. Similarly,  \cite{DBLP:conf/iclr/YangZXK20} shows that for most Atari games, the value function has a low-rank structure. By leveraging the knowledge that the value function has a low-rank structure, \cite{DBLP:conf/iclr/YangZXK20} designs a method with low-rank matrix reconstruction to enforce the low-rank structure of the trained Q network through modifying the fitting target of the Q network ($y^{(i)}$ in eq. \ref{eq:dqn-loss}), which results in better performance on more than half of the Atari games.

\section{Related Work}
\subsection{Applications of reinforcement learning on power grids}
Reinforcement learning (RL) trains an agent how to take actions in an environment to maximize the pre-defined reward \cite{li2017deep}. During recent years, researchers implemented RL algorithms on electrical grids to facilitate grid management, predict future demand, improve grid resilience under extreme situations, etc. They simulated and trained smart grids on virtual environments with different RL algorithms, including Q-learning, DQN, SARSA, DDPG, and more advanced RL algorithms \cite{zhang2018review}. For example, Lu and Hong \cite{lu2019incentive} proposed a real-time incentive-driven demand response algorithm based on RL for smart grids. Kim et. al. \cite{kim2014dynamic} developed a dynamic pricing system to improve the efficiency of grid management with Q-learning. Marino et. al. \cite{marino2016building} utilized long short term memory (LSTM) algorithms for load prediction to improve grid flexibility. François-Lavet et. al. \cite{franccois2016deep} designed a CNN architecture for large-scale energy management and operational control. In addition to these studies, there are considerable number of papers related to RL applications on electrical grids. Based on these related studies, we determined the most prevalent algorithms to be selected when conducting our comparative research and quantitative metrics for grid performance evaluation.

\subsection{Low-rank regularization methods}
Regularization terms have been widely adopted in neural network training to prevent over-fitting. For example, L1 and L2 regularization terms discussed in \cite{DBLP:conf/nips/KroghH91}. Furthermore, when the target we aim to learn is assumed to be low-rank, low-rank regularization (LRR) methods are widely used to enforce the low-rank property of the learned function, which have achieved great success in many data analysis tasks. For example, Hu et al. \cite{DBLP:journals/nn/Hu00L21} provided a comprehensive review of the applications of LRR methods.

Recently, some researchers studied the regularization of value functions for RL. For example, Piché et. al. \cite{DBLP:journals/corr/abs-2106-02613} proposed a functional regularization approach to increase the stability of training of value-based RL without a target network, which leads to significant improvements on sample efficiency and performance across a range of Atari and simulated robotics environments. In addition, Grau-Moya et. al. \cite{DBLP:conf/iclr/Grau-MoyaLV19} proposed a value-based RL algorithm that uses mutual-information regularization to optimize a prior action distribution for better performance and exploration.

While LRR is commonly used in various machine learning tasks to enforce the low-rank structure of the trained model, to the best of our knowledge, adopting LRR to value-based deep RL (i.e., DQN) has not been studied before. As a result, our study is a pioneering work on this topic.

\section{Methodology}
\subsection{Reinforcement learning on Grid2Op with different observation and action spaces}

\subsubsection{Description of the Environment}

This section describes the experimental environment we used in detail, including the virtual grid framework, Grid2Op, and two scenarios for simulation: a simpler scenario with only 5 substations, rte\_case5\_example, and a more complex scenario with 14 substations, rte\_case14\_realistic.

\textit{Grid2Op}:    Grid2Op\footnote{https://grid2op.readthedocs.io/en/latest/} is an open-source Python package for modern electrical grids simulation \cite{marot2021learning}. It is adopted as the virtual environment for many well-known grid design competitions such as L2RPN \footnote{https://www.epri.com/l2rpn}. In this study, we implemented RL algorithms on Grid2Op and modified its built-in action space and observation space.

\textit{rte\_case5\_example}:    This scenario is a simple example of an electrical grid with only 5 substations. We used it as a baseline scenario.

\begin{figure}[h]
  \centering
  \includegraphics[width=\linewidth]{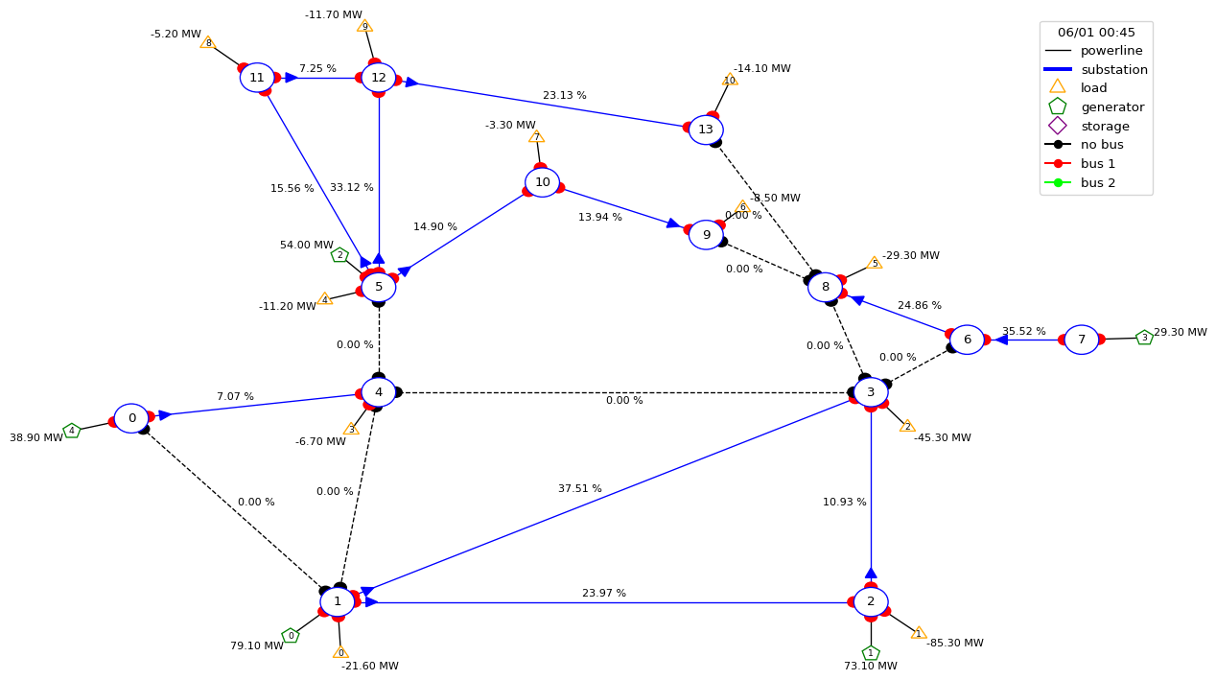}
  \caption{rte\_case14\_realistic Scenario with 14 Substations}
\end{figure}

\textit{rte\_case14\_realistic}:     As a more complicated and realistic scenario, rte\_case14\_realistic contains 14 substations. Fig.1 illustrates the scenario at some timestamp with high load demand.

\subsubsection{Action Space}
An action means an operation (e.g. make a bus change) the grid can take at each step, the action space contains all possible operations of the grid. In this study, we selected three action spaces for experiment: \textit{Topology}, \textit{Powerline Set}, and \textit{Topology Set}. The specification of each action space is shown in Table 1.

\begin{table*}
  \caption{Action Spaces}
  \label{tab:commands}
  \begin{tabular}{ccc|ccc}
  \toprule \multicolumn{3}{c}{Action Spaces} & \multicolumn{1}{c}{} \\
    \toprule
    Topology & Powerline Set & Topology Set & Meaning\\
    \midrule
    set\_line\_status & set\_line\_status & set\_line\_status & reconnect/disconnect the powerline or do nothing\\
    change\_line\_status & & & change the powerline from connected to disconnected or vice versa\\
    set\_bus & & set\_bus & reconnect the object to a different bus or do nothing\\
    change\_bus & & & change the bus that the object is connected to\\
    \bottomrule
  \end{tabular}
\end{table*}

\begin{table*}
  \caption{Observations Spaces}
  \label{tab:commands}
  \begin{tabular}{cc|c}
  \toprule \multicolumn{2}{c}{Observation Spaces} & \multicolumn{1}{c}{} \\
    \toprule
    Complete & Essential & Meaning\\
    \midrule
    gen\_p & gen\_p & the active power production in MW of each power generator\\
    gen\_q...  & & the reactive production in MVar of each power generator\\
    load\_p & load\_p & the active power consumption in MW of each power load\\
    load\_q... & & the reactive consumption in MVar of each power load\\
    p\_or & p\_or & the active power flow in MW at the origin end of each powerline \\
    a\_or & a\_or & the current power flow in MW at the origin end of each powerline \\
    q\_or... & & the reactive power flow in MVar at the origin end of each powerline \\
    p\_ex & p\_ex & the active power flow in MW at the extremity end of each powerline \\
    a\_ex & a\_ex & the current power flow in MW at the extremity end of each powerline \\
    q\_ex... & & the reactive power flow in MVar at the extremity end of each powerline\\
    rho & rho & the capacity of each powerline without unit \\
    line\_status & line\_status & the status of each powerline as a vector \\
    timestep\_overflow & timestep\_overflow & the number of time steps since a powerline is in overflow\\
    topo\_vect & topo\_vect & the bus that each object is connected to as a vector\\
    ... &  & ... \\
    \bottomrule
  \end{tabular}
\end{table*}

\textit{Topology}: Four actions are available in this action spcace. They are set\_line\_status, change\_line\_status, set\_bus, and change\_bus. 

\begin{itemize}
    \item set\_line\_status has three possible values, +1, 0, or -1, which mean reconnect the powerline, do nothing, or disconnect the powerline. 
    
    \item change\_line\_status has two possible values, true or false, which mean change the status of the powerline (i.e. from connected to disconnected or from disconnected to connected) or do nothing.
    
    \item set\_bus has four possible values, +2, +1, 0, or -1, which mean connect the object to bus 2, connect the object to bus 1, do nothing, or disconnect the object. 
    
    \item change\_bus has two possible values, true or false, which mean change the bus of the object (i.e. from bus 1 to bus 2 or from bus 2 to bus 1) or do nothing.
\end{itemize}

\textit{Powerline Set}: It is a subset of \textit{Topology} that contains only one action, set\_line\_status.

\textit{Topology Set}: It is a subset of \textit{Topology} that contains two actions, set\_line\_status and set\_bus.

\subsubsection{Observation Space}

An observation means the accurate information of a feature that is observed without noise (e.g. the power on a bus). The observation space contains all information without noise. In this study, we selected two observation spaces for experiment: \textit{Complete} and \textit{Essential}. The specification of each observation space is shown in Table 2.

\textit{Complete}: This observation space contains all possible observations.

\textit{Essential}: It is a subset of \textit{Complete} that only contains the most essential observations.

\begin{itemize}
    \item gen\_p/load\_p: the active power production/consumption in MW of each power generator/load.
    
    \item p\_or/a\_or: the active/current power flow in MW at the origin end of each powerline.
    
    \item p\_ex/a\_ex: the active/current power flow in MW at the extremity end of each powerline.
    
    \item rho: the capacity of each powerline without unit. 
    
    \item line\_status: the status of each powerline as a vector (e.g. "True" in the $i^{th}$ position means the $i^{th}$ powerline is connected).
    
    \item timestep\_overflow: the number of time steps since a powerline is in overflow.
    
    \item topo\_vect: the bus that each object is connected to as a vector (e.g. "1" in the $i^{th}$ position means the $i^{th}$ object is connected to bus 1). 
    
\end{itemize}

\subsubsection{Evaluation Metrics}
We used three evaluation metrics for algorithm performance evaluation in this study, \textit{Time-based}, \textit{Event-based}, and \textit{Economic-based}. 

\textit{Time-based}: Time-based metrics evaluate the time lapse from response to recovery. For example, from the starting point $t=1$ when a contingency occurs to a recovery state at $t=\tau \in [1,H]$, where $H$ indicates the horizon of the episode. One common instance of the time duration metric is the recovery duration ($TD_r$) defined by,

\begin{equation}\label{TD}
TD_r=\sum_{g\in \mathcal{G}_r} \min({\tau_g,H}),
\end{equation}

Where $\mathcal{G}_r$ is a predefined partition of the graph $G$, and $\tau_g$ is the recovery time of the subgraph $g\in \mathcal{G}_r$. Note that $\tau_g$ can be larger than the episode length $H$, in which $H$ is used in the summation (refer to the min operator).

\textit{Event-based}: Event-based metrics examine the severity of a contingency in the aftermath. We proposed two event-based metrics. Let $d_{sc,j}$ and $d_{ac,j}$ denote the scheduled and actual power demands of load $j\in V$, respectively, and let $c_i\in\{0,1\}$ denote the connectivity status of powerline $i\in E$ as either connected ($c_i=1$) or disconnected ($c_i=0$), then load satisfaction (LS) and line connectivity (LC) are given by,

\begin{equation}\label{LS}
LS =\frac{\sum_{j\in V}  d_{ac,j}}{\sum_{j\in V} d_{sc,j}} ,
\end{equation}
\begin{equation}
LC=\frac{\sum_{i\in E} c_i}{|E|},
\end{equation}

Where $|E|$ is the number of powerlines of the grid. Note that the above metrics can be calculated at every time step within the episode (e.g. instantaneous).

\textit{Economic-based}: Economic-based metrics capture the additional costs incurred due to a response that entails market reactions (e.g. power generation redispatch), and consequently, economic compensations. Note that topological actions are typically cheaper as the grid asset is within the operator's control \cite{fisher2008optimal}. Let $p_{sc,k}$ and $p_{ac,k}$ denote the scheduled and actual power production of generator $k$, respectively, and $c_{re}$ be the unit cost of redispatch (uniform among generators). Under the same setting as the Robustness Track in L2RPN NeurIPS 2020, we defined operational cost (OC) as follows,

\begin{equation}
OC=c_{re} \sum_k{|p_{sc,k}-p_{ac,k}|},
\end{equation}

We formulated the resilience problem as a standard Markov decision process $(S,A,\mathcal{T},R,H,\gamma)$, defined by a state space $S$, an action space $A$, a transition function $\mathcal{T}:S\times A\to \mu(S)$, where $\mu(S)$ is a probability measure over $S$, and a reward function $R:S\times A\times S\to \mathbb{R}$ \cite{sutton2018reinforcement}. While $\mathcal{T}$ is unknown, the agent has the access to transition samples by either interacting with the environment directly or through offline experience replay in the form of quadruples $(s_t,a_t,s_{t}',r_t)$, where $s_t,a_t$, and $r_t$ denote the state, action, and reward at time $t$, respectively, and $s_t'$ is the next state. The agent is tasked to find a stationary policy $\pi:S\to \mu(A)$ that maximizes the expected cumulative return, $\mathbb{E}[\sum_{t=1}^H\gamma^t r_t]$, where $\gamma\in[0,1]$ is the discount factor. Here $\pi$ can be either deterministic or stochastic, with $\mu(A)$ is a Dirac delta function or a generic probability measure over action space $A$, respectively.

\subsection{Low-rank regularization on deep Q-learning}

\cite{DBLP:conf/iclr/YangZXK20} shows that for most of the Atari games, the value function has low-rank structures and, by enforcing a low-rank structure of the value network, the performance of DQN agents can be improved on more than half of the Atari games. Inspired by the result, we believed that using regularization on the loss function to enforce the value network to have a low-rank structure can achieve a similar result. In this case, the loss function in equation \ref{eq:dqn-loss} becomes,

\begin{equation} \label{eq:dqn-loss-with-reg}
\sum_{i=1}^{|\mathcal{B}|}(y^{(i)}-Q(s_{t}^{(i)},a_{t}^{(i)};\theta))^2 + \lambda  R(Q_{\mathcal{B}})
\end{equation}

Where $\lambda$ is a hyper-parameter to control how heavily the value network is regularized, $Q_{\mathcal{B}} \in \mathbb{R}^{|\mathcal{B}| \times |\mathcal{A}|}$ is the output of the value network given $\{(s_{t}^{(i)}\}_{i=1}^{|\mathcal{B}|}$ as the input, and $R(Q_{\mathcal{B}})$ as the rank of $Q_{\mathcal{B}}$. Since the rank of a matrix is computationally expensive and indifferentiable, $R(.)$ is often replaced as the convex envelope of the rank (e.g. nuclear norm). Denote $f(.) \in \mathcal{S} \rightarrow \mathbb{R}^{|\mathcal{A}|}$ as the value function that inputs a state and outputs the expected future reward after taking each action in the action space as a vector form,

\begin{equation} \label{eq:q-vector}
Q_{\mathcal{B}} =
\begin{bmatrix}
f(s_{t}^{(1)})^{T}\\
f(s_{t}^{(2)})^{T}\\
...\\
f(s_{t}^{(|\mathcal{B}|)})^{T}
\end{bmatrix}
\end{equation}

Denoting $\sigma_i$ as the $i$-th singular value of matrix $Q_{\mathcal{B}}$, some commonly used regularization terms are listed in Table 3.

\begin{table}[h]
\centering
\caption{
    Common Low-rank Regularization Terms 
}
    \begin{tabular}{lc}
    \hline 
    \textbf{Type} & $\mathcal{R}(Q_{\mathcal{B}})$ \\ 
    \hline 
    Nuclear norm & $\sum_{i=1}^k \sigma_i$ \\
    Log nuclear norm \cite{peng2015subspace} & $\sum_{i=1}^k log(\sigma_i+1)$ \\
    Elastic-net regularization \cite{DBLP:conf/cvpr/KimLO15} & $\sum_{i=1}^k (\sigma_i + \gamma {\sigma_i^2)}$   \\
    Schatten-$p$ norm \cite{DBLP:conf/aaai/NieHD12, DBLP:journals/jmlr/MohanF12} & $\sum_{i=1}^k \sigma_i^p$ \\
    Truncated nuclear norm \cite{DBLP:journals/pami/HuZYLH13} & $\sum_{i=r+1}^k \sigma_i$ \\
    Partial sum nuclear norm \cite{DBLP:conf/iccv/OhKTBK13} & $\sum_{i=r+1}^k \sigma_i$ \\
    Weighted nuclear norm \cite{DBLP:conf/cvpr/GuZZF14,DBLP:journals/ijcv/GuXMZFZ17,DBLP:conf/aaai/ZhongXLLC15} & $\sum_{i=r+1}^k w_i \sigma_i$ \\
    \hline 
    \end{tabular}
    \label{tab:regularization-type}
\end{table}

\section{Experiments}
\subsection{A resilient power grid simulator}

We evaluated the resilient Grid2Op simulator using existing RL algorithms, including deep Q-learning (DQN), double dueling deep Q-Learning (DDQN), and AlphaDeesp. Besides, we verified the environment behaves in the same way when no resilience setting is activated. Furthermore, we designed a contingency event to verify that the RL agent is able to learn a policy from interacting with the environment when the resilience setting is involved. 

The experimental scenario is a standard IEEE-14 bus system with 14 substations, 5 generators, and 11 loads, rte-case14-realistic. Under traditional settings, the termination of an episode should be consistent with the length of a dataset; in our experiment, we curtailed the episodic length to 100 steps since it is well enough for evaluating the resilience response. In addition, we used an artificial event that cuts down several powerlines happens before the training and returns the already-attacked state to the agent as the initial state of the episode. Furthermore, we set the agent with only essential observations, including the power generation and consumption at each substation, power flow at the origin and extreme of each powerline, status and thermal limit of each powerline, number of time steps before a powerline disconnects because of overflow, cool-down time for each component, and the network's topology vector. Finally, we set the agent to perform topology and powerline-related actions.

We used a double-headed fully-connected neural network to predict the advantage function and value function for each state-action pair. The hidden layers have $(2O,O,896,512)$ neurons, where $O$ represents the observation size. Both the advantage head and value head are fully connected networks with a 384-neuron hidden layer. Between each two layers is a leaky ReLU layer with $\alpha=0.01$. The Q-value is computed by adding the output of the advantage head and value head. The optimizer is Adam optimizer with a learning rate of $1\times 10^{-4}$ that decays at a rate of 0.95 every 1,000 steps. Four consecutive observations are combined as the input for the network.

\begin{figure}
\centering
\includegraphics[width=0.5\textwidth]{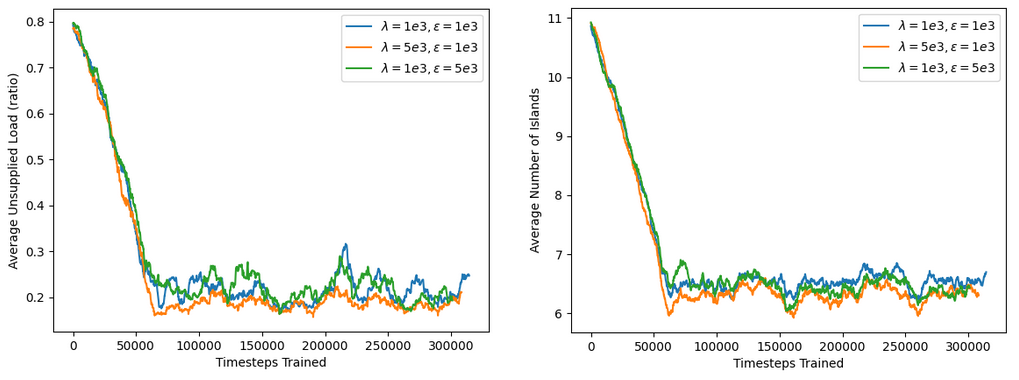}
\caption{Test results of DDQN with X-axis as the number of episode. Y-axis represents the average unsupplied load on the left figure, and average number of islands on the right figure.}
\label{res1}
\end{figure}

\begin{figure}
\centering
\includegraphics[width=0.5\textwidth]{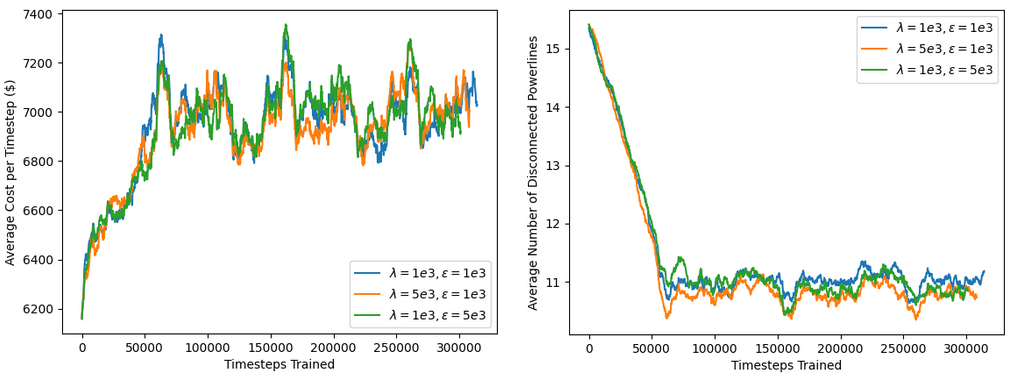}
\caption{Test results of DDQN with X-axis as the number of episode. Y-axis represents the average cost per time step on the left figure, and average number of disconnected powerlines on the right figure.}
\label{res2}
\end{figure}

Fig. \ref{res1} and Fig. \ref{res2} show the training performance of a DDQN agent. We trained the agent using $\lambda=\{1e3, 5e3\}$ and $\epsilon=\{1e3, 5e3\}$. We verified that with all hyperparameter combinations, the agent is able to learn a policy and improve its training performance. Although we found the agents couldn't converge to an optimal policy when we checked the actions taken against a test-time contingency, the results demonstrate that the resilience environment setting supports learning. Additionally, Table \ref{tab:ddqn_result} shows the test-time results from all metrics using a DDQN agent trained with $\lambda=5e3$ and $\epsilon=1e3$.

\begin{table}[h]
\centering
\caption{
    Test-time Results from a DDQN Agent with $\lambda=5e3$ and $\epsilon=1e3$
}
    \begin{tabular}{lcc}
    \hline
     Metrics & Mean & Standard Deviation \\ 
    \hline
     Steps survived & 99.00 & 0.00\\
     Cost & 6944.88 & 2840.12 \\
     Number of islands & 6.03 & 1.49 \\
     Unsupplied load & 0.18 & 0.18 \\
     Broken powerlines & 10.30 & 4.75\\
     \hline
    \end{tabular}
    \label{tab:ddqn_result}
\end{table}

\subsection{Reinforcement learning on Grid2Op with different action and observation spaces}

Table \ref{tab:space} shows RL algorithms' highest average time steps survived and reward smoothed over 100 episodes with respect to different action and observation spaces. 

We found that reducing the action space to powerline-set is highly effective for improving the performance, as it might increase the reward and time steps survived two-folds. Reducing the observation space is also positive for improving the performance, but the improvement is often limited. 

The highest reward and time steps survived for both algorithms are from using the \textit{Essential} observation space and \textit{Powerline Set} action space. Note that an agent that does nothing might survive for about 400 time steps, so we can confidently say that the powerline-set agent we trained successfully learned a policy for dealing with network failure.

\begin{table*}[h]
\centering
\caption{
    Action and Observation Spaces: A Comparative Study
}
    \begin{tabular}{lcccc}
    \hline 
    Algorithms & Observation Spaces & Action Spaces & Time Steps Survived & Reward \\ 
    \hline 
    DDQN & Complete & Powerline Set & 783.1 & $8.7\times 10^5$\\
    DDQN & Complete & Topology & 335.3 & $4.1\times 10^5$\\
    DDQN & Complete & Topology Set & 364.9 & $4.18 \times 10^5$\\
    DDQN & Essential & Powerline Set & \textbf{789.9} & $9.0\times 10^5$\\
    DDQN & Essential & Topology & 669.4 & $7.5\times 10^5$\\
    DDQN & Essential & Topology Set & 631.4 & $7.1\times 10^5$\\
    SliceRDQN & Complete & Powerline Set & 857.3 & $9.5\times 10^5$\\
    SliceRDQN & Complete & Topology & 433.5 & $5.0\times 10^5$\\
    SliceRDQN & Complete & Topology Set & 424.9 & $5.0\times 10^5$\\
    SliceRDQN & Essential & Powerline Set & \textbf{1027.0} & $1.0\times 10^6$\\
    SliceRDQN & Essential & Topology Set & 450.3 & $2.1\times 10^5$\\
    \hline 
    
    \end{tabular}
    \label{tab:space}
\end{table*}

\subsection{Low-rank regularization on deep Q-learning}

We implemented the proposed low-rank regularization method on Grid2op rte\_case5\_example scenario with double deep Q-learning (DQN). We evaluated the agent's performance based on the number of time steps survived and mean reward. We tested our proposed method with different levels of regularization by varying the $\lambda$ term in Equation \ref{eq:dqn-loss-with-reg}. When $\lambda=0$, the agent is a regular DQN agent without regularization.

The experimental results are shown in Table \ref{tab:lrr-exp-results}. For each value of $\lambda$, we trained 14 different agents with random seeds for 200k time steps. Furthermore, for each trained agent, we evaluated its performance by calculating the average performance 50 times. Therefore, each value in Table \ref{tab:lrr-exp-results} is the average of 700 data points (14 agents times 50 times of evaluation).

The agents' performance increases significantly from baseline by adding a small level of regularization ($\lambda = 10^{-8}$). The performance has a huge drop at $\lambda = 10^{-7}$ and increases until another peak at $\lambda = 10^{-3}$. This implies that with a proper value of the regularization term $\lambda$, our proposed method can increase the performance of the agent significantly. 

\section{Conclusion}

In this study, we first presented our findings of implementing a resilient power grid simulator based on Grid2Op. Our modification enables simulation after the grid is divided into multiple sub-networks. We verified the correctness of our modification by testing the environment transitions and demonstrated that the simulator supports RL using the same interface as its predecessor. In the future, we plan to test more contingency events and develop better RL algorithms that can improve power grid resilience.  

We then presented the comparative study of changing both observation and action spaces. Through experiments using value-based algorithms including DDQN and SliceRDQN, we concluded that reducing the size of action space is highly effective for training.  

Finally, we proposed a low-rank regularization method and demonstrated that, with proper value of the $\lambda$ term, our proposed method can increase the performance of the agents significantly. In the future, more experiments should be performed in different environments to test if the proposed method can generalize well.

\begin{table}[h]
\centering
\caption{
    Performance of DQN Agents with Different Levels of Low-rank Regularization
}
    \begin{tabular}{lcc}
    \hline
     $\lambda$ & Reward & Times Steps Survived  \\ 
    \hline
     $0$ (baseline) & \textit{5900} & \textit{814} \\ 
     $10^{-8}$ & \textbf{8412} & \textbf{1176} \\ 
     $10^{-7}$ & 5073 & 703 \\ 
     $10^{-6}$ & 5816 & 805 \\ 
     $10^{-5}$ & 6504 & 882 \\ 
     $10^{-4}$ & 6724 & 936 \\ 
     $10^{-3}$ & \textbf{8285} & \textbf{1120}
     \\ 
     $10^{-2}$ & 6508 & 920 \\ 
     \hline
    \end{tabular}
    \label{tab:lrr-exp-results}
\end{table}

\bibliographystyle{ACM-Reference-Format}
\bibliography{sample-base}
\end{document}